\documentclass{article}

\usepackage{microtype}
\usepackage{graphicx}
\usepackage{svg}
\usepackage{subfigure}
\usepackage{booktabs} 
\usepackage{xcolor}
\usepackage[frozencache=true,cachedir=minted-cache]{minted}

\usepackage{hyperref}
\usepackage{float}

\usepackage[accepted]{icml2023}

\usepackage{amsmath}
\usepackage{amssymb}
\usepackage{mathtools}
\usepackage{amsthm}
\usepackage{xfrac}

\usepackage[capitalize,noabbrev]{cleveref}

\theoremstyle{plain}

\theoremstyle{definition}

\theoremstyle{remark}

\usepackage[textsize=tiny]{todonotes}
\usepackage[normalem]{ulem}

\icmltitlerunning{OpenDiLoCo}

\begin{document}

\twocolumn[
\icmltitle{OpenDiLoCo: An Open-Source Framework for Globally Distributed Low-Communication Training}




\begin{icmlauthorlist}
\icmlauthor{Sami Jaghouar}{comp}
\icmlauthor{Jack Min Ong}{comp}
\icmlauthor{Johannes Hagemann}{comp}
\end{icmlauthorlist}

\icmlaffiliation{comp}{Prime Intellect, Inc.}

\icmlcorrespondingauthor{Johannes Hagemann}{johannes@primeintellect.ai}

\icmlkeywords{Distributed Training, Decentralized AI, Local-SGD, DiLoCo}

\vskip 0.3in
]



\printAffiliationsAndNotice{} 

\begin{abstract}
OpenDiLoCo is an open-source implementation and replication of the Distributed Low-Communication (DiLoCo) training method for large language models. We provide a reproducible implementation of the DiLoCo experiments, offering it within a scalable, decentralized training framework using the Hivemind library. We demonstrate its effectiveness by training a model across two continents and three countries, while maintaining 90-95\% compute utilization. Additionally, we conduct ablations studies focusing on the algorithm's compute efficiency, scalability in the number of workers and show that its gradients can be all-reduced using FP16 without any performance degradation. Furthermore, we scale OpenDiLoCo to $3\times$ the size of the original work, demonstrating its effectiveness for billion parameter models.

\end{abstract}

\section{Introduction}
\label{submission}
Large language models (LLMs) have revolutionized numerous applications of machine learning, yet training these models requires substantial computational resources typically concentrated in a single, well-connected cluster to efficiently parallelize workloads for distributed model training~\cite{hagemann2023efficient}.
Novel approaches, such as DiLoCo by \citeauthor{douillard2023diloco}, address these challenges by enabling efficient training across multiple, poorly connected devices.
Their approach dramatically reduces the need for frequent communication, making it feasible to train LLMs on a global scale.\\
We reproduce DiLoCo's results in an open manner and implement them in a real-world setting using the Hivemind library~\cite{hivemind}, showcasing its applications and analyzing its compute efficiency.
In summary, the contributions of our work are as follows:

\begin{itemize}
    \item \textbf{Reproduction and Scaling of DiLoCo Experiments:} We replicate the DiLoCo experiments for a large language model pre-training and validate their results in a reproducible manner. We also successfully extend the DiLoCo experiments to the billion-parameter model scale.
    \item \textbf{Open-Source Implementation:} We provide implementations of DiLoCo built on top of the Hivemind library alongside a concise 180-line PyTorch version, significantly lowering the barrier for performing decentralized training. Our framework enables single DiLoCo workers to scale to hundreds of machines through our integration with PyTorch FSDP.
    \item \textbf{Global Decentralized Training:} We demonstrate our approach in a real-world decentralized training setting executed across two continents and three countries, achieving 90-95\% compute utilization.
    \item \textbf{Analytical Insights and Ablations:} We conduct an ablation study of DiLoCo, focusing on the algorithm's scalability in the number of workers and compute efficiency. We also demonstrate that DiLoCo pseudo gradients can be effectively all-reduced using FP16 without any performance degradation.
\end{itemize}

We publish the full data of our experiments, the Hivemind as well as the PyTorch distributed training code implementation of OpenDiLoCo on GitHub at \href{https://github.com/PrimeIntellect-ai/OpenDiLoCo}{github.com/PrimeIntellect-ai/OpenDiLoCo}.

\section{Implementation}

DiLoCo is a local SGD algorithm ~\cite{stich2019local} that leverages two distinct optimization processes: an inner optimizer and an outer optimizer.
The inner optimizer, AdamW  ~\cite{DBLP:journals/corr/abs-1711-05101}, performs local updates on individual workers, while the outer optimizer, SGD with Nesterov momentum ~\cite{Nesterov1983AMF}, synchronizes the workers using pseudo-gradients calculated by subtracting the locally updated weight $\theta(t+h)$ from the original weight $\theta(t)$.

This local SGD approach significantly reduces the frequency of communication (up to 500 times), thus lowering the bandwidth requirements for distributed training.

\paragraph{General Implementation Details}

Our implementation of DiLoCo instantiates the two optimizers (inner and outer) and creates two copies of the model: the main model $\theta(t+h)$, which will be updated by the inner optimizer, and a copy of the original weights, $\theta(t)$, which is needed to compute the pseudo-gradient. The inner optimizer is called at the end of each step, while the outer optimizer is called periodically. Both of our implementations compute the pseudo-gradients manually and store them in FP32 inside the PyTorch gradient buffer (within $param.grad$) of the model. Further experiments show that the pseudo gradient can be stored and all-reduced in FP16 without noticeable performance hit. See \autoref{fp16_all_reduce}.

In mixed precision training~\cite{DBLP:journals/corr/abs-1710-03740} with FP16, a gradient scaler is used to improve the dynamic range of the gradients while avoiding underflow and overflow. The gradient scaler should be called during the inner optimization step but not during the outer one because the pseudo-gradients are calculated manually in FP32.

We offer two open-source DiLoCo implementations, one reference implementation using \texttt{torch.distributed}
and an implementation built using the Hivemind library for a more practical decentralized training setting.

\paragraph{Implementation with \texttt{torch.distributed} }

The following details our PyTorch implementation, utilizing the \texttt{torch.distributed} package with NCCL for the communication backend.

\usemintedstyle{borland}
\renewcommand{\theFancyVerbLine}{\textcolor[RGB]{0,0,0}{\small \arabic{FancyVerbLine}}}
\begin{figure}[H]
\begin{minted}[
fontfamily=courier,
fontsize=\fontsize{7pt}{7pt},
xleftmargin=8pt, 
numbersep=4pt, 
linenos, 
frame=lines,
baselinestretch=1.5]{python}
for batch, step in enumerate(train_loader):
    ... # loss calculation
    inner_optimizer.step()
    if real_step % local_steps == 0:
        for old_param, param in \
                zip(original_params, model.parameters()):         
                
            param.grad = old_param - param.data
            dist.all_reduce(
                tensor=param.grad, 
                op=dist.ReduceOp.AVG
            )
            param.data = old_param 
        outer_optimizer.step()
    original_params = [
        p.detach().clone() for p in model.parameters()
    ]
\end{minted}
\vspace{-4mm}
 \caption{\textbf{Pseudo-Code for Outer Optimizer in OpenDiLoCo.}}
\label{fig:torch-distributed-pseudo-code}
\end{figure}

We highlight the most important outer optimization part in \autoref{fig:torch-distributed-pseudo-code}.

Due to the use of a dual optimizer setup and the calculation of pseudo-gradients, this implementation requires custom training code, making it incompatible out of the box with popular training scripts from Hugging Face or PyTorch Lightning.
The communication in this implementation also uses the NCCL backend which cannot communicate across networks using NAT, preventing its use over the internet.
Our second implementation using Hivemind alleviates both of these issues.

\paragraph{Hivemind Implementation}

The following implementation is built on top of the Hivemind framework~\footnote{\href{https://github.com/learning-at-home/hivemind}{github.com/learning-at-home/hivemind}}. Instead of using \texttt{torch.distributed} for the worker communication, Hivemind utilizes a distributed hash table (DHT) spread across each worker to communicate metadata and synchronize them. This DHT is implemented using the open-source libp2p project~\footnote{\href{https://github.com/libp2p/libp2p}{github.com/libp2p/libp2p}}. Hivemind provides an optimized all-reduce algorithm designed for execution on a pool of poorly connected workers.

\usemintedstyle{borland}
\renewcommand{\theFancyVerbLine}{\textcolor[RGB]{0,0,0}{\small \arabic{FancyVerbLine}}}
\begin{figure}[H]
\begin{minted}[
fontfamily=courier,
fontsize=\fontsize{7pt}{7pt},
xleftmargin=8pt, 
numbersep=4pt, 
linenos, 
frame=lines,
baselinestretch=1.5]{python}
from hivemind.dht.dht import DHT
from open_diloco import DiLoCoOptimizer

optimizer = DiLoCoOptimizer(
    bs, # batch size
    ls, # learning rate scheduler
    DHT(), # distributed hash table for coordination
    i_opt, # inner optimizer
    o_opt, # outer optimizer
    m.params() # model parameters
)

for batch in train_dataloader:

    model(batch).loss.backward()
    optimizer.step() 
    # the outer step, including peer synchronization 
    # and communication, is triggered automatically 
    # after all local steps
    optimizer.zero_grad()
\end{minted}
\vspace{-4mm}
 \caption{\textbf{OpenDiLoCo - Hivemind API.}}
\label{fig:opendiloco-hivemind-api}
\end{figure}

Our integration with Hivemind enables a real-world decentralized training setup for DiLoCo, making many of its inherent properties usable, such as:
\begin{itemize}
    \item \textbf{On/Off ramping of resources:} The amount of available compute may not be constant, with new devices and clusters coming and going.
    \item \textbf{Fault tolerance:} For decentralized training, some devices may be less reliable than others. Through Hivemind's fault-tolerant training, a device could become unavailable at any time without stopping the training process.
    \item \textbf{Peer-to-Peer:} There is no master node. All communication is done in a peer-to-peer fashion.
\end{itemize}

Unlike the \texttt{torch.distributed} implementation, our Hivemind implementation wraps both optimizers into a single optimizer class, making it compatible with popular training codebases that assume a single optimizer, such as the Hugging Face Trainer. This allows for the use of OpenDiLoCo via a simple Hivemind-compatible API by instantiating a customizable DiLoCoOptimizer, as shown in \autoref{fig:opendiloco-hivemind-api}.

Additionally, our custom implementation allows to combine both Hivemind and PyTorch FSDP ~\cite{zhao2023pytorchfsdpexperiencesscaling}, enabling to scale single DiLoCo workers to multiple nodes or whole clusters.

\section{Experiments}

\paragraph{Replication Experiment Setup}
Our OpenDiLoCo replication experiment setup largely follows the main experiments from \citeauthor{douillard2023diloco}.
We conduct various experiments using a model with 150 million parameters on a language modeling task using the C4 dataset~\cite{c4}.
The hyperparameters are consistent with DiLoCo across experiments: an inner learning rate of $4e^{-4}$, $1{,}000$ warm-up steps, $0.1$ weight decay, a batch size of $512$, a sequence length of $1{,}024$, a learning rate for the Nesterov outer optimizer of $0.7$, and Nesterov momentum of $0.9$.
Similarly, we run the experiments for a total of $88{,}000$ steps.\\
The one difference in our experiment setup is that we choose the Llama~\citep{llama} model architecture for our experiments, due to its recent popularity, while the original DiLoCo authors used the Chinchilla architecture~\cite{chinchilla}.
These two architectures are generally quite similar but have slight differences.
For instance, Llama uses the SwiGLU activation function~\cite{swiglu} for the MLP and has a dimension of $\frac23 4d$ instead of $4d$.
For more details about the model configuration, see \autoref{appendix:model-configuration}.

In addition to the DiLoCo experiments, we conduct experiments with a varying number of workers to analyze if diminishing returns occur before reaching the 8 workers reported in the DiLoCo work and to generally measure the FLOP efficiency of the algorithm.

We also run experiments in a real-world decentralized training setup, training across workers from three different countries simultaneously.

\begin{table*}
\centering
\resizebox{0.75\linewidth}{!}{%
\begin{tabular}{@{}l|cccc@{}}
\toprule
Model & Communication & Time & Compute \& Data & Perplexity \\
\midrule
Baseline, no replica, from scratch & 0 & $1\times$ & $1\times$ & 16.17 \\
Baseline, $8\times$ batch size with DP & $8\times N$ & $1\times$ & $8\times$ & 13.68 \\
\textbf{DiLoCo, 8 replicas, 500 local steps} & $8\times \frac{N}{H}$ & $1\times$ & $8\times$  & \textbf{13.73} \\
\bottomrule
\end{tabular}
}
\caption{\textbf{Final Evaluation Perplexity Comparison}: We compare our two baselines \textit{vs} DiLoCo with 8 replicas for a 150 million parameter model pre-training across their communication cost, time spent, compute \& data used and final perplexity after $88{,}000$ steps, similar to ~\citeauthor{douillard2023diloco}.
For the same time and amount of compute, we can compare the second baseline and DiLoCo.
The former communicates gradients at each time step ($N$ total steps), while DiLoCo communicates $H=500$ times less.}
\label{tab:baselines}
\end{table*}

Our baselines also follow a similar setup as \citeauthor{douillard2023diloco}.
We use two baselines: the first is a weak baseline that runs without DiLoCo and without replicas for $88{,}000$ steps.
The second is a stronger baseline, which uses an $8\times$ larger batch size with data parallelism, maintaining a similar compute budget as our DiLoCo experiment but with significantly larger communication requirements.

\paragraph{Main Results}
\label{main-results}

\begin{figure}
  \centering
  \includegraphics[width=\linewidth]{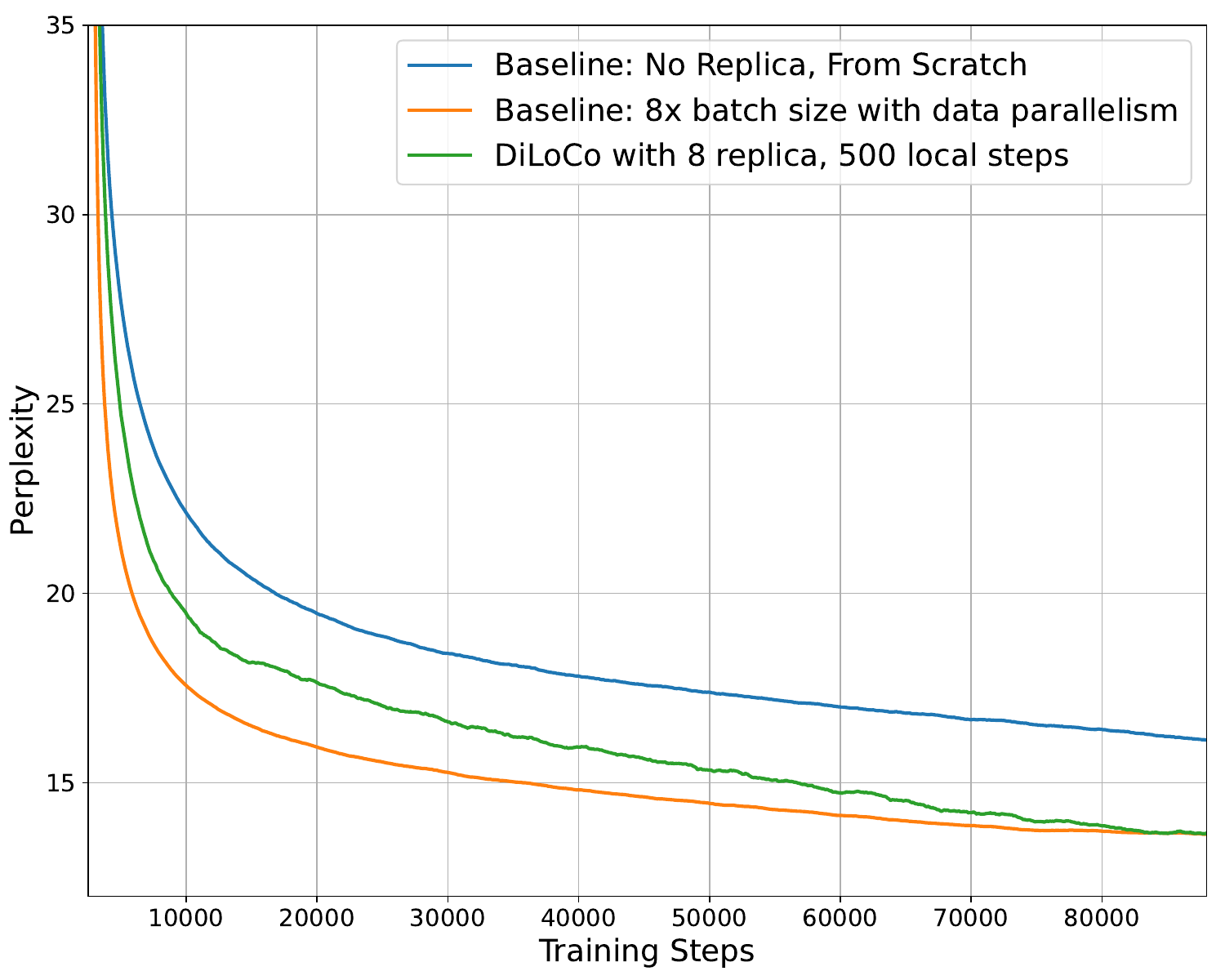}
  \vspace{-6mm}
  \caption{\textbf{Main result:} 150 million parameter Llama model pre-training with 8 DiLoCo workers yields significantly lower perplexity than the baseline without DiLoCo, and even compared to the baseline using 8 times larger batch size with the same compute budget, while communicating 500 times less.
  }
  \label{fig:diloco-8-worker-llama-150m-experiment}
\end{figure}

\autoref{fig:diloco-8-worker-llama-150m-experiment} shows our main experimental results. It demonstrates that DiLoCo with 8 replicas significantly outperforms the baseline without any replicas and matches the performance of the stronger baseline with $500\times$ larger communication requirements, as indicated by the final perplexity results in \autoref{tab:baselines}.
These findings are consistent with the main experimental results of \citeauthor{douillard2023diloco}.
One noticeable difference is that in \citeauthor{douillard2023diloco}'s experiments, the DiLoCo run is already approaching and surpassing the stronger baseline at around $64{,}000$ steps, while our DiLoCo training run only starts to exactly match the performance of the strong baseline at the end of the training at $88{,}000$ steps. This difference might be due to the fact that their experiments start from a checkpoint with $24{,}000$ pre-training steps, while ours start from scratch.

\begin{figure}
  \centering
  \includegraphics[width=\linewidth]{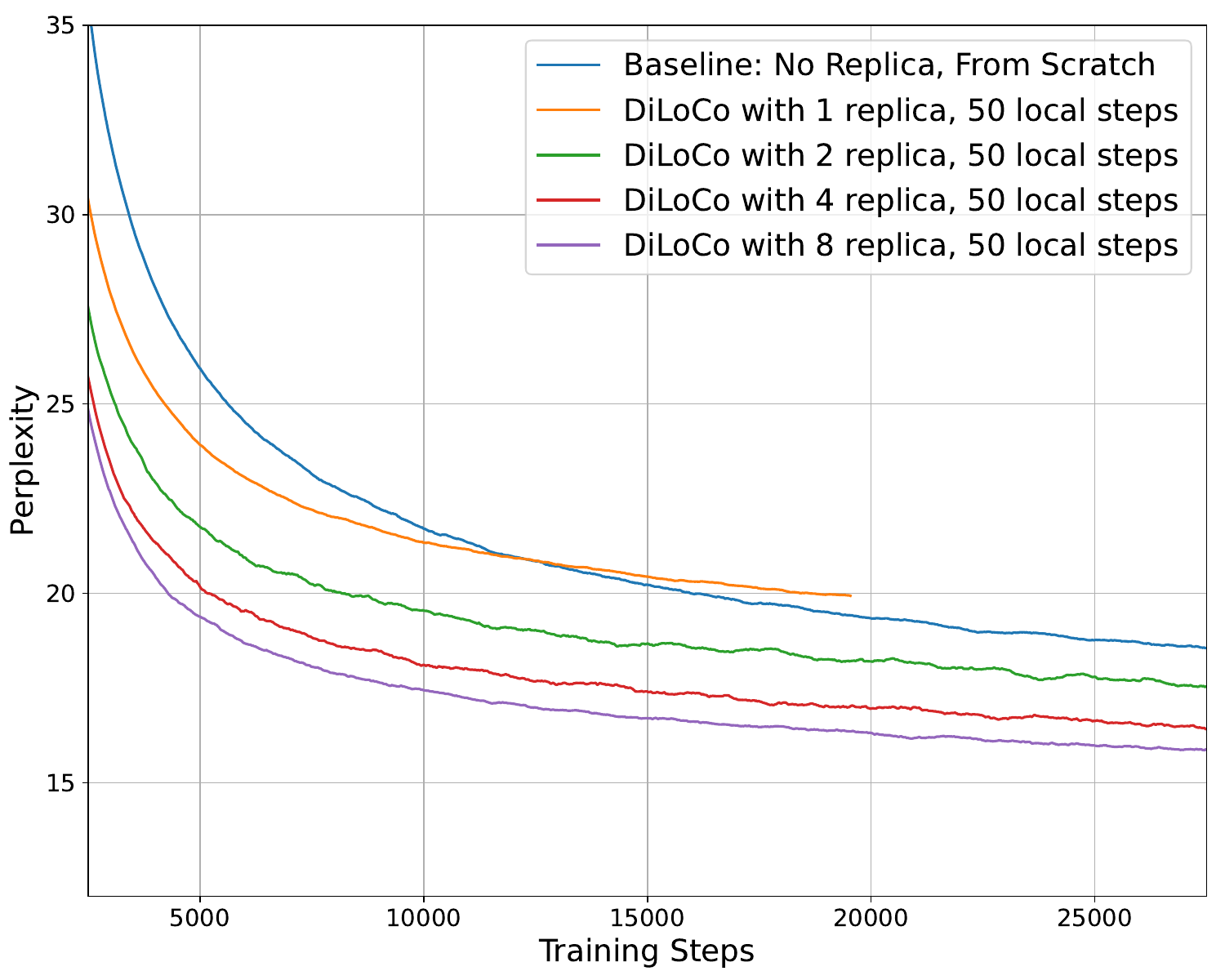}
  \vspace{-6mm}
    \caption{\textbf{Ablation Study on the Number of Workers in DiLoCo:} Performance comparison of DiLoCo with different numbers of workers and 50 local steps against the baseline without DiLoCo. Due to compute constraints, these ablation experiments were not extended to $88{,}000$ steps like the other experiments.}
  \label{fig:diloco-num-of-worker-llama-150m-experiment-ablation}
\end{figure}

\paragraph{Number of Worker and FLOP Efficiency Ablation}
To determine the compute efficiency of DiLoCo, we conduct an ablation study on the number of workers, as shown in \autoref{fig:diloco-num-of-worker-llama-150m-experiment-ablation}.
These experiments are set up identically to our main experiment, with the only difference being a reduction in the local step size from 500 to 50.

Our results demonstrate a steady improvement in perplexity as the number of workers in DiLoCo increases.

\begin{figure}[!htbp]
  \centering
  \includegraphics[width=\linewidth]{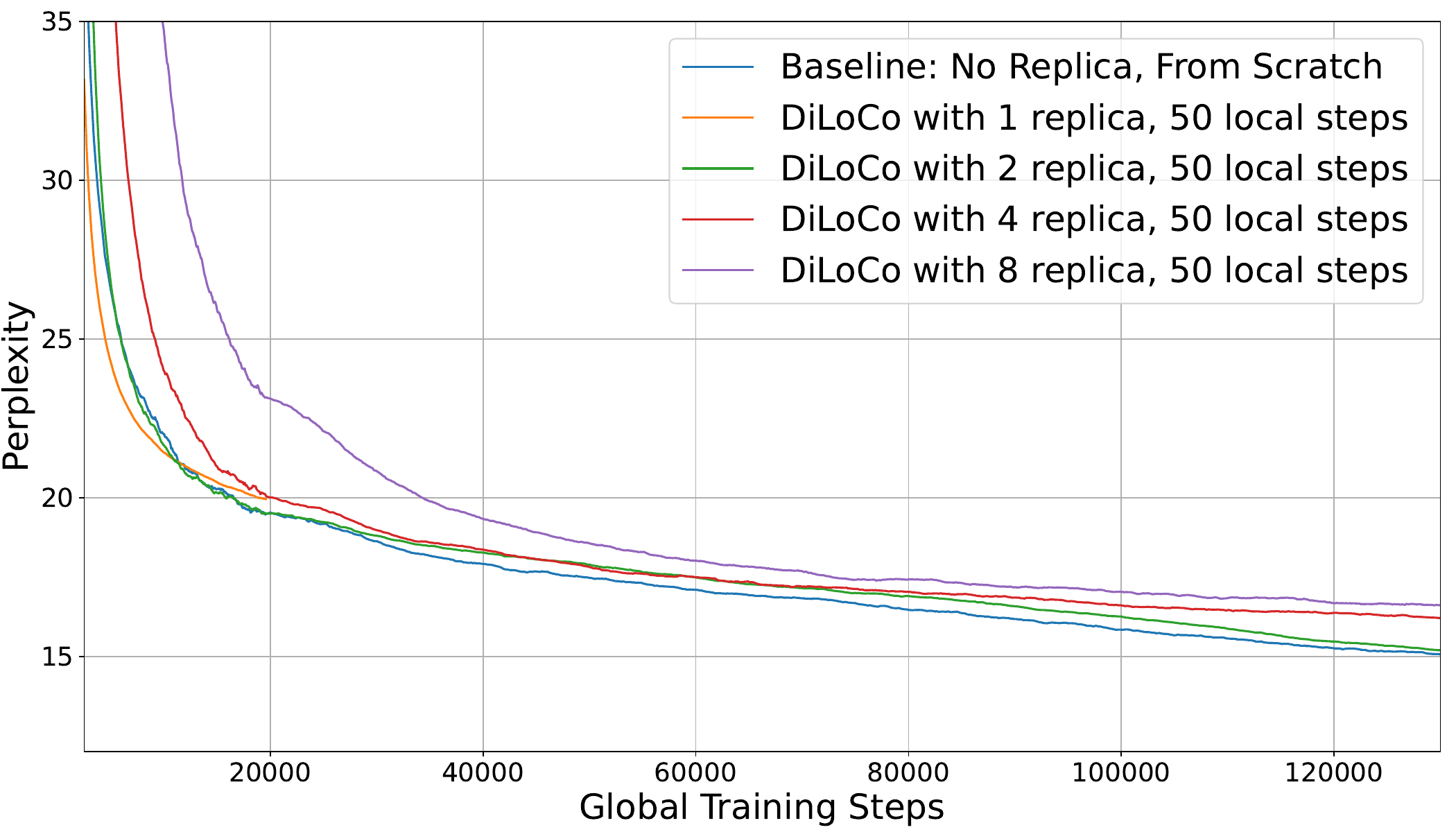}
    \caption{\textbf{Ablation Study on FLOP Efficiency Relative to Number of Workers in DiLoCo:} This figure compares the performance of DiLoCo with different numbers of workers and 50 local steps against the baseline without DiLoCo. The x-axis shows the global steps instead of local steps, providing a better approximation of DiLoCo's FLOP efficiency by comparing the total amount of compute spent on the model.}
  \label{fig:diloco-num-of-worker-llama-150m-experiment-ablation-global-steps}
\end{figure}

Furthermore, \autoref{fig:diloco-num-of-worker-llama-150m-experiment-ablation-global-steps} presents the same ablation as \autoref{fig:diloco-num-of-worker-llama-150m-experiment-ablation}, but with the x-axis representing global steps instead of local steps.
This provides a more accurate approximation of DiLoCo's FLOP efficiency by comparing the total compute spent on the model.
These results reinforce our previous observation: DiLoCo with more than one worker is initially not as compute efficient as the same number of global steps on a single machine or when using Distributed Data Parallel training.
DiLoCo may only achieve comparable FLOP efficiency after a large number of steps due to slower initial convergence, as shown in our main experiment in \autoref{fig:diloco-8-worker-llama-150m-experiment}.

\paragraph{Practical Usage} 
According to our main experimental results in \autoref{fig:diloco-8-worker-llama-150m-experiment}, eight DiLoCo workers yield a final perplexity comparable to that of DDP after $88{,}000$ steps. However, training for only $44{,}000$ steps with eight workers results in a significantly worse performing model than DDP with the same number of global steps, making four DiLoCo workers a more efficient choice in this case. Our interpretation suggests that while training with eight DiLoCo workers ultimately results in a stronger model, increasing the number of workers does not accelerate the initial convergence phase as data parallelism would.

\phantomsection
\label{fp16_all_reduce}

\paragraph{All-Reduce in FP16}

Our main experiments perform the all-reduce operation of the pseudo gradients in FP32, following the original methodology outlined in the DiLoCo paper. We repeated the DiLoCo experiment, this time using FP16 for the pseudo gradient.

\begin{figure}
  \centering
  \includegraphics[width=\linewidth]{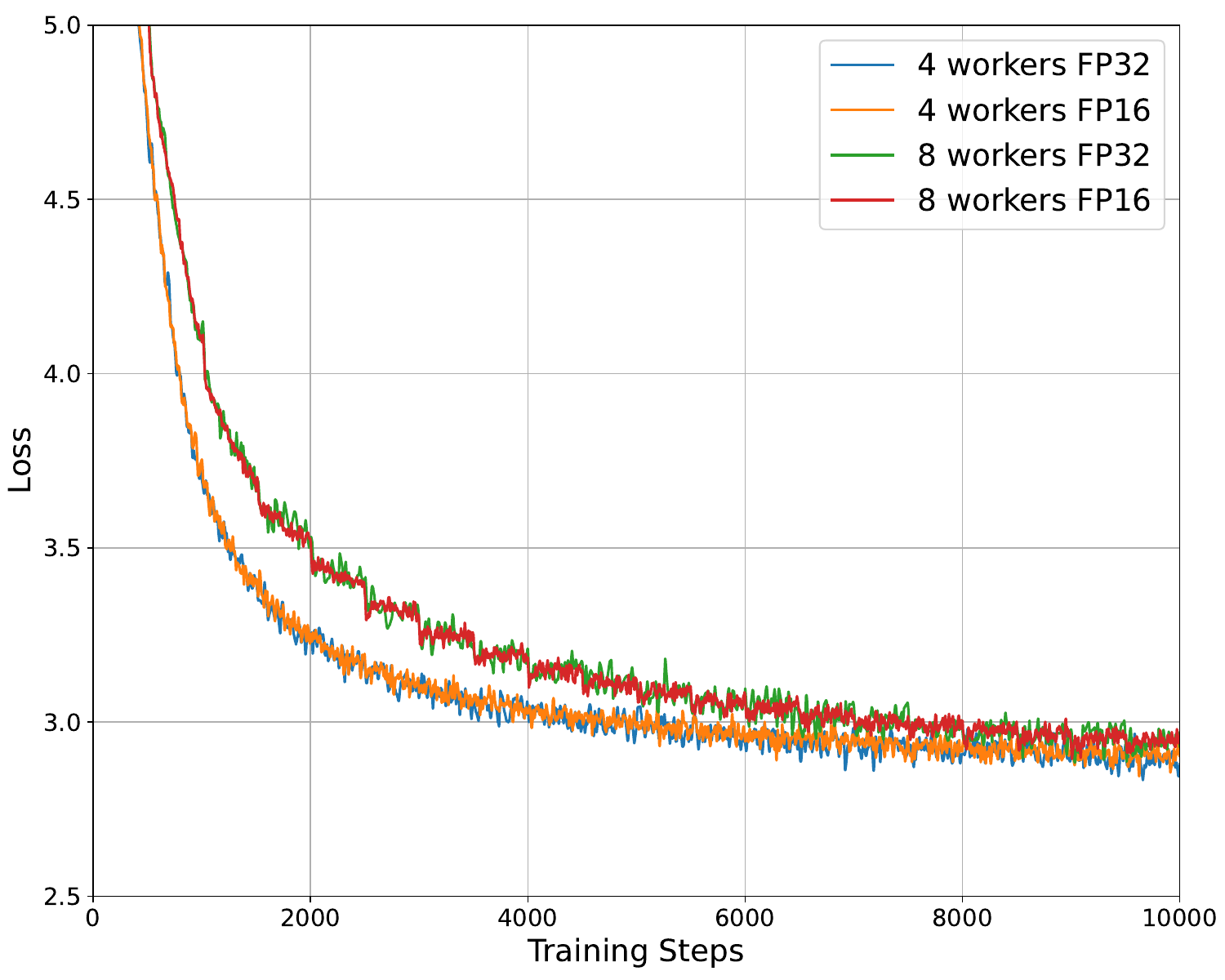}
  \vspace{-6mm}
    \caption{\textbf{FP16 vs FP32 All-Reduce Ablation:} The first group is 4 workers and 50 local steps, the second group is 8 workers and 500 local steps.}
  \label{fig:diloco-fp16}
\end{figure}

\autoref{fig:diloco-fp16} shows there is no noticeable impact on performance both with $8$ workers and $500$ local steps and $4$ workers and $50$ local steps, indicating that FP16 all-reduce is effective for use with DiLoCo and can halve the communication time required for the all-reduce operation.

\paragraph{Scaling DiLoCo to Billion Parameter Models}

The original DiLoCo paper demonstrated the efficacy of the method up to model sizes of 400 million parameters. We expand on this and test the scalability of DiLoCo to larger models sizes by pre-training a 1.1 billion parameter model.

We adopt the same hyperparameters as TinyLlama~\cite{zhang2024tinyllamaopensourcesmalllanguage}, employing a model with 1.1B parameters, a learning rate of $4e^{-4}$ and a batch size of $2048$. We conduct our experiment with four workers and an outer learning rate of $0.7$. We experiment with two DiLoCo runs for this model size: the first with 500 local steps, as in our experiments in \autoref{fig:diloco-8-worker-llama-150m-experiment}, and the second with 125 local steps. Since the batch size of this training run is 4 times larger than the batch size of our main experiment, the second run with 125 local steps effectively has the same number of tokens per outer step on each DiLoCo worker as the main experiment with 500 local steps.

We compare our results against two baselines: a weak baseline without DiLoCo and without replicas, and a stronger baseline using a $4\times$ larger batch size with data parallelism, maintaining a similar compute budget as the DiLoCo experiment.

Our baselines are trained using PyTorch FSDP~\cite{zhao2023pytorchfsdpexperiencesscaling} with the hybrid sharding strategy on two co-located nodes, each equipped with eight H100 GPUs. For our DiLoCo experiments, each of the four DiLoCo workers operates on individual nodes with eight H100 GPUs. Intra-node communication is handled by FSDP using the NCCL backend to leverage fast interconnect speeds, while Hivemind manages inter-node, low-bandwidth communication. We use the no-shard strategy in FSDP to avoid incompatibility between the Hivemind state averager and FSDP Sharded DTensors.

\begin{figure}  
  \centering
  \includegraphics[width=\linewidth]{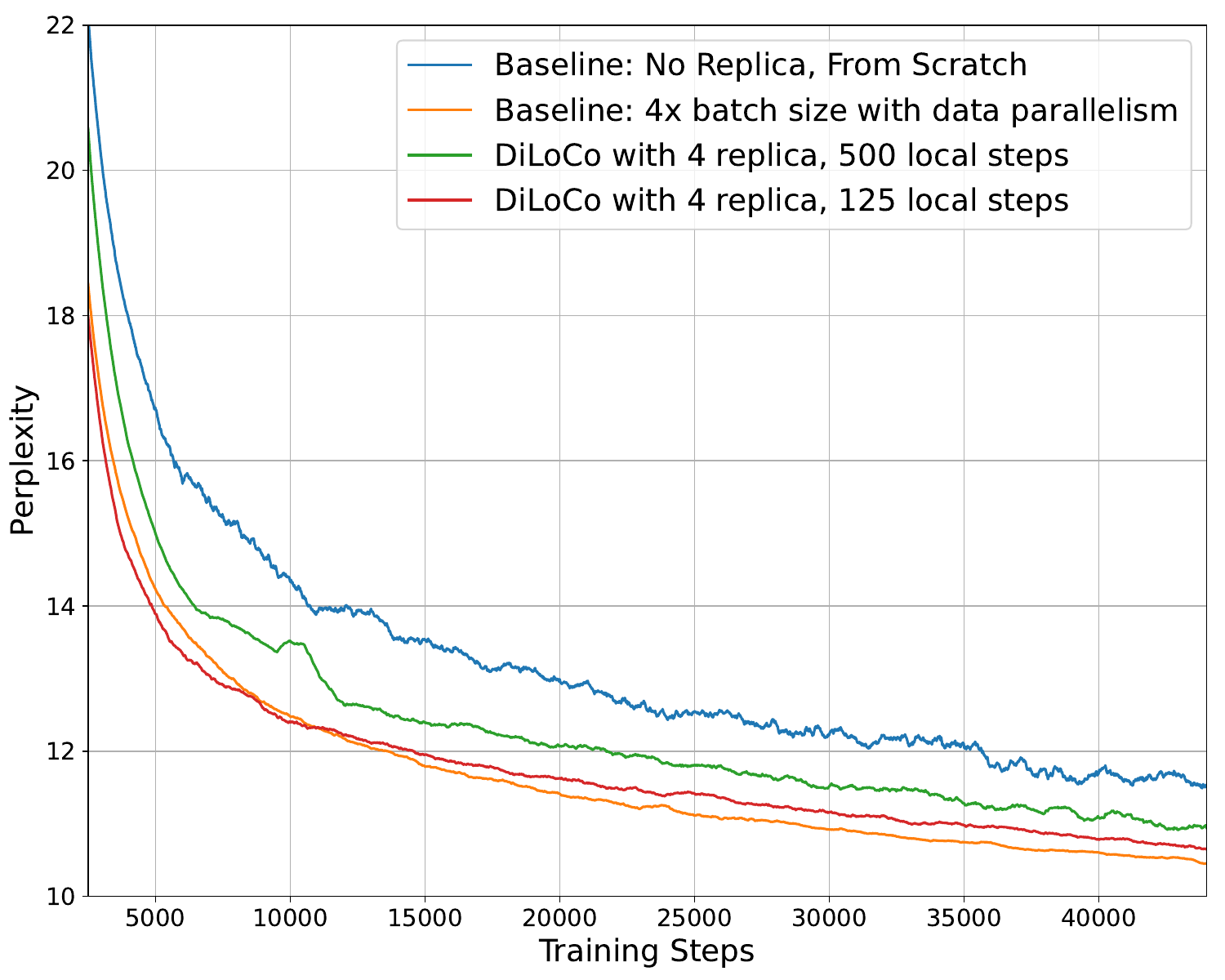}
  \vspace{-6mm}
    \caption{\textbf{1.1B Scaling Experiment}. Comparing a 1.1B training with OpenDiLoCo with 4 workers syncing every 500 local steps and every 125 local steps against the two baselines.}
  \label{fig:diloco-1b}
\end{figure}

As depicted in \autoref{fig:diloco-1b}, both DiLoCo experiments significantly outperform the weaker baseline. However, only the OpenDiLoCo run with 125 local steps nearly matches the performance of the stronger baseline with the same compute budget, while communicating 125 times less. The final perplexity difference between the 4 worker DiLoCo run with 125 local steps and the stronger baseline is $0.24$ as show in \autoref{tab:baselines_1b}

\begin{table*}
\centering
\resizebox{0.75\linewidth}{!}{%
\begin{tabular}{@{}l|cccc@{}}
\toprule
Model & Communication & Time & Compute \& Data & Perplexity \\
\midrule
Baseline, no replica, from scratch & 0 & $1\times$ & $1\times$ & 11.85 \\
Baseline, $4\times$ batch size with DP & $4\times N$ & $1\times$ & $4\times$ & 10.52 \\
\textbf{DiLoCo, 4 replicas, 125 local steps} & $8\times \sfrac{N}{H}$ & $1\times$ & $4\times$  & \textbf{10.76} \\
\textbf{DiLoCo, 4 replicas, 500 local steps} & $8\times \sfrac{N}{H}$ & $1\times$ & $4\times$  & \textbf{11.14} \\
\bottomrule
\end{tabular}
}
\caption{\textbf{Final Perplexity Comparison}: We compare our two baselines \textit{vs} DiLoCo with 4 replicas for a 1.1B parameter model pre-training across their communication cost, time spent, compute \& data used and final perplexity after $44{,}000$ steps.
For the same time and amount of compute, we can compare the second baseline and DiLoCo.
The former communicates gradients at each time step ($N$ total steps), while DiLoCo communicates $H$ times less (with $H=125$ or  $H=500$) }
\label{tab:baselines_1b}
\end{table*}

We propose a hypothesis for why the OpenDiLoCo run with 500 local steps underperforms compared to the stronger baseline: \\
In our initial experiment with the 150m model, we run for a total of $88,000$ steps. For the scaled-up 1.1B parameter experiment, we limit it to $44,000$ steps because of the $4\times$ larger batch size. This means that for the same number of training tokens, the DiLoCo synchronization happens only a quarter of the time as often in the 1.1B experiment compared to the 150m experiment. This makes the well-performing 125 local step experiment a better comparison. However, even in the 500 local steps DiLoCo run, we observe faster convergence in the later stages of training, gradually catching up to the stronger baseline.

While we demonstrate that DiLoCo works at the billion parameter scale, we believe that further work is needed to make it effective with even larger batch sizes and more local steps.

\paragraph{Globally Distributed Training Setting}

\begin{figure}
  \centering
  \includegraphics[width=0.75\linewidth]{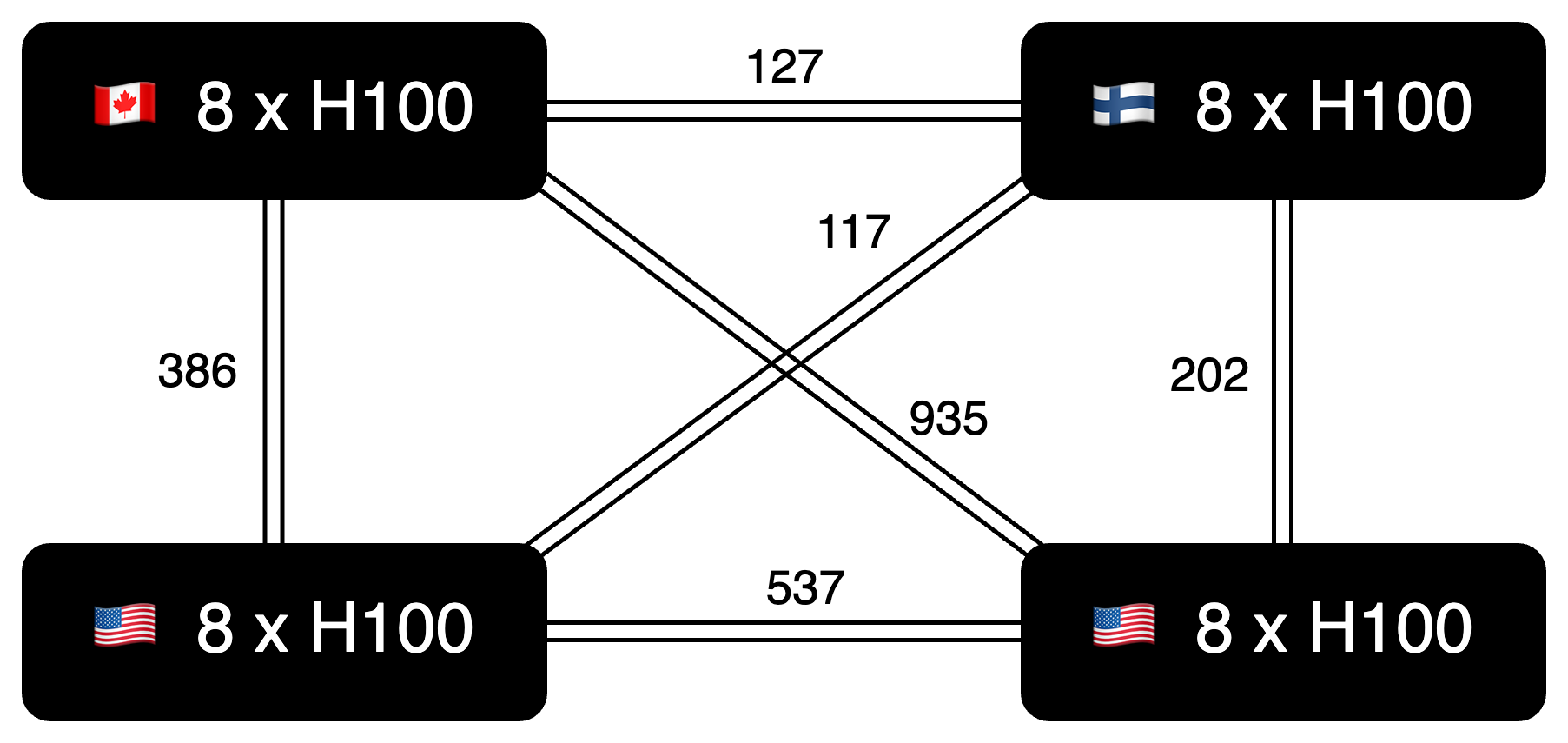}
  \caption{\textbf{Network Bandwidth between Workers:} Average bidirectional network bandwidth between the four different workers in our decentralized training setup (\textbf{in Mbit/s}). The GPUs are located in three different countries and hosted by different cloud providers: Canada (Hyperstack); Finland (DataCrunch); United States, Texas (Voltage Park); and United States, Delaware (Runpod). Measured using the iperf package~\footnote{iperf package: \url{https://packages.ubuntu.com/jammy/iperf3}}.}
  \label{fig:connection-speeds-decentralized-training}
\end{figure}

To showcase the functionality of decentralized training with OpenDiLoCo executed across different continents, we utilize our Hivemind implementation. We use four DiLoCo workers, each with eight H100 GPUs, located in Canada, Finland, and two different states within the United States. \autoref{fig:connection-speeds-decentralized-training} shows the network bandwidth between the workers, which varies between 127 to 935 Mbit/s. We train our 1.1B parameter model with 500 local steps, as in our scaling experiment. The gradients are all-reduced in FP16.

Through the large number of local steps, the four workers run independently for around 67.5 minutes before communicating for gradient averaging. For the outer optimizer step, our experiment shows an average all-reduce time between the workers of 300 seconds. 

Additionally, we observe variations in the training speed between our different cloud instances. Although all workers have the same GPU type, we could not control for configuration variables such as the number of CPU cores and the amount of RAM, which led to slightly different training times for the 500 inner steps. \\
Nevertheless, due to the significant reduction in communication time, the all-reduce bottleneck only accounts for 6.9\% of the training time, minimally impacting the overall training speed.
Additional training time is spent idling by the fastest worker in our scenario. In future work, we will address this issue by exploring DiLoCo in an asynchronous setting, as done by ~\citeauthor{asyncdiloco}.

\section{Conclusion}
We successfully reproduce the main experiment results of DiLoCo, scale the method to $3\times$ the parameter size of the original work and demonstrate its application in a real-world decentralized training setting.
We train a large language model using our OpenDiLoCo implementation across 2 continents and 3 countries and achieve 90-95\% compute utilization through the low-communication training approach.

We show that DiLoCo exhibits strong performance with two or four replicas, opening up practical applications. However, while scaling DiLoCo to more than eight workers is a promising research direction for enabling effective, low-communication training across globally distributed GPUs, our ablation study shows using eight workers does not yet match the computational efficiency of Distributed Data Parallel (DDP) training when running for a shorter amount of steps. 

For future work, more compute-efficient methods need to be developed for decentralized training, which also improve the scalability to support a significantly larger number of distributed workers.
More sophisticated model merging techniques could be used to improve stability and convergence speed.
On top of that, compute idle time could be reduced by implementing methods that perform the weight averaging communication asynchronously, interleaving them with the computation for the next outer optimizer update.\\
Additional efforts will be directed towards scaling OpenDiLoCo to test the algorithm's scaling behavior on even larger model sizes, further enhancing its applicability and efficiency in real-world scenarios.

\section*{Acknowledgements}

We want to thank Max Ryabinin for his guidance and help with the Hivemind library. His insights have been very helpful for our project.

We would also like to thank Arthur Douillard for his work on DiLoCo and for helping us figure out the details of reproducing the original experiments.

\bibliography{example_paper}

\begin{thebibliography}{14}
\providecommand{\natexlab}[1]{#1}
\providecommand{\url}[1]{\texttt{#1}}
\expandafter\ifx\csname urlstyle\endcsname\relax
  \providecommand{\doi}[1]{doi: #1}\else
  \providecommand{\doi}{doi: \begingroup \urlstyle{rm}\Url}\fi

\bibitem[Douillard et~al.(2023)Douillard, Feng, Rusu, Chhaparia, Donchev, Kuncoro, Ranzato, Szlam, and Shen]{douillard2023diloco}
Douillard, A., Feng, Q., Rusu, A.~A., Chhaparia, R., Donchev, Y., Kuncoro, A., Ranzato, M., Szlam, A., and Shen, J.
\newblock Diloco: Distributed low-communication training of language models, 2023.

\bibitem[Hagemann et~al.(2023)Hagemann, Weinbach, Dobler, Schall, and de~Melo]{hagemann2023efficient}
Hagemann, J., Weinbach, S., Dobler, K., Schall, M., and de~Melo, G.
\newblock Efficient parallelization layouts for large-scale distributed model training, 2023.

\bibitem[Hoffmann et~al.(2022)Hoffmann, Borgeaud, Mensch, Buchatskaya, Cai, Rutherford, de~Las~Casas, Hendricks, Welbl, Clark, Hennigan, Noland, Millican, van~den Driessche, Damoc, Guy, Osindero, Simonyan, Elsen, Rae, Vinyals, and Sifre]{chinchilla}
Hoffmann, J., Borgeaud, S., Mensch, A., Buchatskaya, E., Cai, T., Rutherford, E., de~Las~Casas, D., Hendricks, L.~A., Welbl, J., Clark, A., Hennigan, T., Noland, E., Millican, K., van~den Driessche, G., Damoc, B., Guy, A., Osindero, S., Simonyan, K., Elsen, E., Rae, J.~W., Vinyals, O., and Sifre, L.
\newblock Training compute-optimal large language models, 2022.

\bibitem[Liu et~al.(2024)Liu, Chhaparia, Douillard, Kale, Rusu, Shen, Szlam, and Ranzato]{asyncdiloco}
Liu, B., Chhaparia, R., Douillard, A., Kale, S., Rusu, A.~A., Shen, J., Szlam, A., and Ranzato, M.
\newblock Asynchronous local-sgd training for language modeling, 2024.

\bibitem[Loshchilov \& Hutter(2017)Loshchilov and Hutter]{DBLP:journals/corr/abs-1711-05101}
Loshchilov, I. and Hutter, F.
\newblock Fixing weight decay regularization in adam.
\newblock \emph{CoRR}, abs/1711.05101, 2017.
\newblock URL \url{http://arxiv.org/abs/1711.05101}.

\bibitem[Micikevicius et~al.(2017)Micikevicius, Narang, Alben, Diamos, Elsen, Garc{\'{\i}}a, Ginsburg, Houston, Kuchaiev, Venkatesh, and Wu]{DBLP:journals/corr/abs-1710-03740}
Micikevicius, P., Narang, S., Alben, J., Diamos, G.~F., Elsen, E., Garc{\'{\i}}a, D., Ginsburg, B., Houston, M., Kuchaiev, O., Venkatesh, G., and Wu, H.
\newblock Mixed precision training.
\newblock \emph{CoRR}, abs/1710.03740, 2017.
\newblock URL \url{http://arxiv.org/abs/1710.03740}.

\bibitem[Nesterov(1983)]{Nesterov1983AMF}
Nesterov, Y.
\newblock A method for solving the convex programming problem with convergence rate $o(1/k^2)$.
\newblock \emph{Proceedings of the USSR Academy of Sciences}, 269:\penalty0 543--547, 1983.
\newblock URL \url{https://api.semanticscholar.org/CorpusID:145918791}.

\bibitem[Raffel et~al.(2019)Raffel, Shazeer, Roberts, Lee, Narang, Matena, Zhou, Li, and Liu]{c4}
Raffel, C., Shazeer, N.~M., Roberts, A., Lee, K., Narang, S., Matena, M., Zhou, Y., Li, W., and Liu, P.~J.
\newblock Exploring the limits of transfer learning with a unified text-to-text transformer.
\newblock \emph{J. Mach. Learn. Res.}, 21:\penalty0 140:1--140:67, 2019.
\newblock URL \url{https://api.semanticscholar.org/CorpusID:204838007}.

\bibitem[Shazeer(2020)]{swiglu}
Shazeer, N.
\newblock Glu variants improve transformer, 2020.

\bibitem[Stich(2019)]{stich2019local}
Stich, S.~U.
\newblock Local sgd converges fast and communicates little, 2019.

\bibitem[team(2020)]{hivemind}
team, L.
\newblock {H}ivemind: a {L}ibrary for {D}ecentralized {D}eep {L}earning.
\newblock \url{https://github.com/learning-at-home/hivemind}, 2020.

\bibitem[Touvron et~al.(2023)Touvron, Lavril, Izacard, Martinet, Lachaux, Lacroix, Rozière, Goyal, Hambro, Azhar, Rodriguez, Joulin, Grave, and Lample]{llama}
Touvron, H., Lavril, T., Izacard, G., Martinet, X., Lachaux, M.-A., Lacroix, T., Rozière, B., Goyal, N., Hambro, E., Azhar, F., Rodriguez, A., Joulin, A., Grave, E., and Lample, G.
\newblock Llama: Open and efficient foundation language models, 2023.

\bibitem[Zhang et~al.(2024)Zhang, Zeng, Wang, and Lu]{zhang2024tinyllamaopensourcesmalllanguage}
Zhang, P., Zeng, G., Wang, T., and Lu, W.
\newblock Tinyllama: An open-source small language model, 2024.
\newblock URL \url{https://arxiv.org/abs/2401.02385}.

\bibitem[Zhao et~al.(2023)Zhao, Gu, Varma, Luo, Huang, Xu, Wright, Shojanazeri, Ott, Shleifer, Desmaison, Balioglu, Damania, Nguyen, Chauhan, Hao, Mathews, and Li]{zhao2023pytorchfsdpexperiencesscaling}
Zhao, Y., Gu, A., Varma, R., Luo, L., Huang, C.-C., Xu, M., Wright, L., Shojanazeri, H., Ott, M., Shleifer, S., Desmaison, A., Balioglu, C., Damania, P., Nguyen, B., Chauhan, G., Hao, Y., Mathews, A., and Li, S.
\newblock Pytorch fsdp: Experiences on scaling fully sharded data parallel, 2023.
\newblock URL \url{https://arxiv.org/abs/2304.11277}.

\end{thebibliography}
\bibliographystyle{icml2023}

\newpage
\appendix
\onecolumn
\section{Model Configuration}
\label{appendix:model-configuration}

\begin{table}[H]
\centering
\begin{tabular}{@{}l|cc@{}}
\toprule
Model Parameters &  150M & 1.1B \\
\midrule
Number of layers & 12 & 22 \\
Hidden dim & $1{,}024$ & $2{,}048$ \\ 
Number of heads & 16 & 32 \\
K/V size & 64 & 64 \\
Vocab size & \multicolumn{2}{c}{$32{,}000$}\\
Inner learning rate (AdamW) & \multicolumn{2}{c}{$4e^{-4}$} \\
Number of warmup steps & \multicolumn{2}{c}{$1{,}000$}\\
Weight decay & \multicolumn{2}{c}{$0.1$}\\
Batch Size & $512$ & $2{,}048$\\
Sequence length & \multicolumn{2}{c}{$1{,}024$}\\
Outer Nesterov learning rate & \multicolumn{2}{c}{$0.7$}\\
Outer Nesterov momentum & \multicolumn{2}{c}{$0.9$}\\
\bottomrule
\end{tabular}
\caption{\textbf{Model Configuration} for the DiLoCo experiments. The models are based on the Llama architecture~\cite{llama}.}
\label{tab:model_config}
\end{table}


\end{document}